\documentclass[10pt,letterpaper]{article}
\usepackage[top=0.85in,left=2.75in,footskip=0.75in]{geometry}

\usepackage{amsmath,amssymb}

\usepackage{changepage}

\usepackage[utf8x]{inputenc}

\usepackage{textcomp,marvosym}

\usepackage{cite}

\usepackage{nameref,hyperref}

\usepackage{microtype}
\DisableLigatures[f]{encoding = *, family = * }

\usepackage[table]{xcolor}

\usepackage{array}

\newcolumntype{+}{!{\vrule width 2pt}}

\newlength\savedwidth

\usepackage{setspace} 
\raggedright
\usepackage[aboveskip=1pt,labelfont=bf,labelsep=period,justification=raggedright,singlelinecheck=off]{caption}

\makeatletter
\renewcommand{\@biblabel}[1]{\quad#1.}
\makeatother

\usepackage{lastpage,fancyhdr,graphicx}
\usepackage{epstopdf}
\fancyheadoffset[L]{2.25in}
\fancyfootoffset[L]{2.25in}
\newcommand{\etal}{et al.\@}

\usepackage{soul}
\usepackage{dsfont}
\usepackage{relsize}
\usepackage{microtype}
\usepackage{xcolor}
\usepackage{booktabs}
\usepackage{algorithm}
\usepackage{algpseudocode}
\usepackage{caption}
\usepackage{subcaption}
\usepackage{amsmath}
\usepackage[inline]{enumitem}
\usepackage{siunitx}

\begin{document}
\vspace*{0.2in}

\begin{flushleft}
{\Large
\textbf\newline{Optimal training of integer-valued neural networks with mixed integer programming}
}
\newline
\\
T{\'o}mas Thorbjarnarson\textsuperscript{1},
Neil Yorke-Smith\textsuperscript{1*}
\\

\bigskip
\textbf{1} Algorithmics group, Faculty of Electrical Engineering, Mathematics and Computer Science, Delft University of Technology, Delft, The Netherlands
\\
\bigskip

* Corresponding author \\ Email: \url{n.yorke-smith@tudelft.nl} (NYS)

\end{flushleft}


\section*{Abstract}
Recent work has shown potential in using Mixed Integer Programming (MIP) solvers to optimize certain aspects of neural networks (NNs).  However the intriguing approach of training NNs with MIP solvers is under-explored.  State-of-the-art-methods to train NNs are typically gradient-based and require significant data, computation on GPUs, and extensive hyper-parameter tuning.  In contrast, training with MIP solvers does not require GPUs or heavy hyper-parameter tuning, but currently cannot handle anything but small amounts of data.  This article builds on recent advances that train binarized NNs using MIP solvers.  We go beyond current work by formulating new MIP models which improve training efficiency and which can train the important class of integer-valued neural networks (INNs).  We provide two novel methods to further the potential significance of using MIP to train NNs.  The first method optimizes the number of neurons in the NN while training.  This reduces the need for deciding on network architecture before training.  The second method addresses the amount of training data which MIP can feasibly handle: we provide a batch training method that dramatically increases the amount of data that MIP solvers can use to train.  
We thus provide a promising step towards using much more data than before when training NNs using MIP models.
Experimental results on two real-world data-limited datasets demonstrate that our approach strongly outperforms the previous state of the art in training NN with MIP, in terms of accuracy, training time and amount of data.
Our methodology is proficient at training NNs when minimal training data is available, and at training with minimal memory requirements -- which is potentially valuable for deploying to low-memory devices.

\section{Introduction}
\label{sec:intro}

Training neural networks (NNs) using gradient-based optimization methods can be tedious.  Hyper-parameters require meticulous and computationally-intensive tuning to reach the best results.  Although state-of-the-art methods use gradient-based optimization, these
methods often require a large number of neurons and immense amounts of data.

In light of the computational demands, a number of recent studies seek to counteract the trend of increasingly-large networks.  For instance, a branch of neural network optimization that intends to reduce the size of networks has gained some traction \cite{DBLP:journals/ijon/HuangSH20,Serra20:compression}.  The motivation of such studies is to decrease the memory needs of NNs and to increase efficiency in training and using them, without degrading the networks' generalization ability.
Further, neural architecture search seeks to automatically obtain for a good configuration of a NN \cite{DBLP:journals/jmlr/ElskenMH19}.

We posit that modelling NNs as mixed integer programs (MIP) and training them with discrete optimization solvers could work well in reduced memory settings.  Recent work shows that this idea is feasible when using minimal data to train the restricted class of binary NNs (BNNs) \cite{DBLP:conf/cp/IcarteICCMB19}.

Training with MIP solvers is not expected to be competitive with gradient-based methods at large scale, as MIP models are currently limited by the amount of data they can use to train.  Instead, we see potential in using MIP solvers to train smaller networks with small batches of data.  Moreover, MIP-based training reduces hyper-parameter tuning considerably: choosing learning rates, momentum, decay, the number of epochs, batch sizes and more becomes unnecessary.  By modelling NNs using MIP and reasonable objective functions, the solver can in principle find a guaranteed optimal solution.  This is theoretically intriguing and of practical advantage, as even after extensive hyper-parameter tuning for gradient-based methods, it can be unclear whether an optimal NN configuration has been reached.

We aim to go beyond existing work that trains BNN by training integer-valued NNs (INNs).  To our knowledge, training INNs with MIP has not yet been researched.  Methods to train INNs can be greatly beneficial as they have greatly reduced memory needs when compared to full-precision 32-bit NNs while still reaching close to state-of-the-art results \cite{DBLP:journals/corr/abs-1808-04752}.  We hypothesize that training INNs with MIP can be a viable alternative to training NNs for low-memory environments, such as on drones or other portable devices.

This article advances over the state-of-art in the literature as follows.
First, we contribute an expandable framework to train INNs using MIP solvers.  This framework provides flexibility such that the user can choose the range of the network's parameters
to adjust the memory usage of the NN.  Our framework can accommodate various loss functions.
Experimental results demonstrate that our approach strongly outperforms the previous state of the art in training NN with MIP, in terms of accuracy, training time and amount of data.

Our second contribution is to provide an optional extension to the MIP models that minimizes the size of the network while training.  Thus our methodology results in networks that are efficient and require very minimal memory.  

We demonstrate on two real-world datasets that our proposed models, solved using the Gurobi MIP solver, can perform comparably to gradient-based methods when minimal data is available.  This can be useful for training on small datasets, and it reduces the need for hyper-parameter tuning and the practical necessity of using GPUs to train NNs.  Going further, our third
contribution is to propose a novel batch training method that is inspired by ensemble training.
Our MIP batch training method exhibits great promise: it can handle larger amounts of data and exploits training in parallel to reduce training time.

The article is structured as follows.
First Section~\ref{sec:approach} motivates and describes 
describe
our approach.
Sections~\ref{sec:expdesign} and~\ref{sec:results} presents 
results of three experiments.
Following,
Section~\ref{sec:rw} positions 
our contribution in the literature.  Section~\ref{sec:conc} concludes  
by identifying future directions.

\section{Modelling approach}
\label{sec:approach}

Prior art, the work of Icarte \etal
\cite{DBLP:conf/cp/IcarteICCMB19}, finds that MIP models are limited by the amount of data they can feasibly use to train NNs.  We aim to increase the amount of data solvers can use to train and to provide models that perform similarly to a gradient descent baseline, given a limited amount of training data.  Further, we aim to capture a range of NN loss functions and non-binary networks.  We then aim to simultaneously train and optimize a network's architecture.

To this end, we begin by proposing three novel NN MIP models that have the same base model but separate objective functions to serve different purposes.  This base NN MIP model captures a multi-layer perceptron with a sign activation function.  The base model which the three models share in common builds on the model from Icarte \etal \cite{DBLP:conf/cp/IcarteICCMB19}.  We modify this base model by replacing the objective function and constraints specific to Icarte's methodology.  We also modify constraints to allow the network's parameters to take larger ranges.  The remaining constraints compute the outputs of each neuron in the NN.

Our base MIP model is described by the following set of constraints.  Note the objective function is discussed below.  The decision variables to optimize are the network's integer weights and biases \eqref{model:param_domains}, continuous connections \eqref{model:conn_domain} and binary activations \eqref{model:bin_domain}.
There are $L$ layers in the network.  The set of layers is denoted as $\mathcal{L} = \{1,...,L\}$. Two additional sets are also denoted to simplify notation; $\mathcal{L}_2 = \{2,...,L\}$ and $\mathcal{L}^{L-1} = \{1,...,L-1\}$. The set $N_{\ell}$ for each layer $\ell$ represents the neurons in the layer. The set $T$ is used to represent all data samples used when training. Thus, $k \in T$ represents a single sample from the training data.

\begin{align}
    & \hat{y}^k_j = \frac{2}{P\cdot (N_{L-1}+1)} \sum_{i \in N_{L - 1}}  c^k_{i L j} + b_{L j}
    & \forall  j \in N_L, k \in T \label{model:yhat} \\
    & (u^k_{\ell j} = 1) \Longrightarrow (\sum_{i \in N_{\ell - 1}} c^k_{i \ell j} + b_{\ell j} \geq 0) & \forall \ell \in \mathcal{L}^{L-1}, j \in N_\ell, k \in T \label{model:bin_pos} \\
    & (u^k_{\ell j} = 0) \Longrightarrow (\sum_{i \in N_{\ell - 1}} c^k_{i \ell j} + b_{\ell j} \leq -\epsilon) & \forall \ell \in \mathcal{L}^{L-1}, j \in N_\ell, k \in T \label{model:bin_neg} \\
    & c^k_{i1j} = x^k_i \cdot w_{i1j} & \forall i \in N_0, j \in N_1, k \in T \label{model:input} \\
    & c^k_{i\ell j} - w^k_{i\ell j} + 2 P \cdot u^k_{(\ell -1)i} \leq 2   {P} & \forall \ell \in \mathcal{L}_2, i \in N_{\ell - 1}, j \in N_\ell, k \in T \label{model:conn1} \\
    & c^k_{i\ell j} + w^k_{i\ell j} - {2 P \cdot} u^k_{(\ell -1)i} \leq 0 & \forall \ell \in \mathcal{L}_2, i \in N_{\ell - 1}, j \in N_\ell, k \in T \label{model:conn2} \\
    & c^k_{i\ell j} - w^k_{i\ell j} - {2 P \cdot} u^k_{(\ell -1)i} \geq -2   {P} & \forall \ell \in \mathcal{L}_2, i \in N_{\ell - 1}, j \in N_\ell, k \in T \label{model:conn3} \\
    & c^k_{i\ell j} + w^k_{i\ell j} + {2 P \cdot} u^k_{(\ell -1)i} \geq 0 & \forall \ell \in \mathcal{L}_2, i \in N_{\ell - 1}, j \in N_\ell, k \in T \label{model:conn4} \\
    & w_{i\ell j}, b_{\ell j} \in \{-P,...,P\} & \forall \ell \in \mathcal{L}, i \in N_{\ell - 1}, j \in N_\ell \label{model:param_domains} \\
    & c^k_{i\ell j} \in \mathds{R} & \forall \ell \in \mathcal{L}, i \in N_{\ell - 1}, j \in N_\ell, k \in T \label{model:conn_domain} \\
    & u^k_{\ell j} \in \{0,1\} & \forall \ell \in \mathcal{L}^{L-1}, j \in N_{\ell}, k \in T \label{model:bin_domain} \\
    & {P \in \mathcal{Z^+}} \label{model:bound_domain}
\end{align}

The variable $w_{i\ell j}$ is the weight of the connection from neuron $i \in N_{\ell -1}$ to neuron $j \in N_\ell$. The variable $b_{\ell j}$ is the bias value of neuron $j \in N_{\ell}$. To model inter-layer calculations for each sample $k$, we use the variable $c^k_{i \ell j}$. The binary variable $u^k_{\ell j}$ models the activation of neuron $j \in N_\ell$ for every sample $k \in T$.  With it we model the sign activation function: if the input to neuron $j \in N_{\ell}$ is negative, $u^k_{\ell j} = 0$ \eqref{model:bin_neg}, otherwise $u^k_{\ell j} = 1$ \eqref{model:bin_pos}.  Subsequently, $c^k_{i \ell j}$ is calculated using $u^k_{\ell j}$.  To properly model the sign function, the values $\{0,1\}$ are mapped to $\{-1,1\}$.  Thus, in following layers, equations (\ref{model:conn1}--\ref{model:conn4}) ensure that $c^k_{i\ell j} = (2u^k_{(\ell -1)i} - 1) \cdot w_{i\ell j}$.  Finally, $\hat{y}^k_j$ models the normalized value in output neuron $j \in N_L$ for sample $k$.
We choose $\epsilon=1 \cdot 10^{-4}$ in equation \eqref{model:bin_neg} to model the inequality in accordance with the variable precision tolerance of the Gurobi MIP solver we will use \cite{Gurobi}.

Equation~\eqref{model:input} models calculations of the NN's first layer using $x_i^k$ to represent neurons in the input layer, $i \in N_0$, for each sample $k$.  Equations (\ref{model:conn1}--\ref{model:conn4}) model the subsequent layers as noted.  Equation \eqref{model:yhat} calculates the values in neurons in the final layer.  The value $\hat{y}^k_j$ therefore represents an encoded predicted label of the network, while $y^k_j$ represents the true encoded label of sample $k$.

All sample labels are encoded using $+1$/$-1$ encoding. Further, the output neurons \eqref{model:yhat} are normalized using a linear approximation method such that all values are approximately between $-1$ and $1$.  Since in a pure MIP model we cannot use non-linear normalization functions, such as softmax or sigmoid, we approximate by a linear normalization.

We next explain our three model variants and their objective functions.  Each model variant adds a different set of constraints and objective to the base model.

\subsection{Model 1: Max-correct}
\label{subsec:max-correct}

Our first
model, \emph{max-correct}, aims to maximize the number of correct predictions of training samples.  It uses a binary variable for each output neuron to denote whether the sample has the correct label.
\begin{align}
    & \max \sum_{k \in T} \sum_{j \in N_L: y^k_j = 1}  p^k_j & \label{max:obj}\\
    & (p^k_j = 1) \Longrightarrow (\hat{y}^k_j \geq 0) & \forall  j \in N_L, k \in T \label{max:o_pos} \\
    & (p^k_j = 0) \Longrightarrow (\hat{y}^k_j \leq -\epsilon) & \forall  j \in N_L, k \in T \label{max:o_neg} \\
    & \sum_{ \forall j \in N_L} p^k_j = 1 & k \in T \label{max:sum_constr} \\
    & p^k_j \in \{0,1\} & \forall j \in N_L, k \in T \label{max:domain}
\end{align}

This model is simple and fast.  It requires only one output neuron per sample to be positive and maximizes the number of positive output neurons that correspond to the correct label.  However, there is little confidence in predictions as they just barely need to be correct.  Therefore, similar samples in the testing dataset may be incorrectly classified.  We will study this empirically in Section~\ref{sec:results}.

\subsection{Model 2: Min-hinge}
\label{subsec:min-hinge}

To increase the confidence of predictions, we propose our second model, \emph{min-hinge}.  This model is inspired by the squared hinge loss \eqref{eq:hinge_loss} that has been shown to perform well when using $+1$/$-1$ encoding for labels \cite{DBLP:journals/corr/JanochaC17}:
\begin{align}
    & L =  \sum_{k \in T} \sum_{j \in N_L} \max\left(0, \frac{1}{2} - \hat{y}^k_j y^k_j\right)^2 \label{eq:hinge_loss}
\end{align}

The squared hinge loss function is non-linear but can be approximated using piecewise linear (PWL) functions.  PWL functions are defined by a number of break-points and lines between the break-points.  By choosing sufficient break-points, the non-linear squared hinge loss function can be approximated.  We can then simply denote the PWL function as $f$ and input the multiplication of our predicted value $\hat{y}^k_j$ \eqref{model:yhat} with the encoded label $y^k_j$ to calculate the loss for a single output neuron for a single sample.

The total loss
is the sum over all output neurons for all samples:
\begin{align}
    & \min \sum_{k \in T} \sum_{j \in N_L}  f( \hat{y}^k_j \cdot y^k_j) & \label{min:obj}
\end{align}

The advantages of this model are that predictions are pushed to be more confident.  The max-correct model's target is to maximize the number of correct predictions.  The min-hinge model also aspires to do so, but additionally aims to make each prediction to be above the margin of $\frac{1}{2}$.

\subsection{Model 3: Sat-margin}
\label{subsec:sat-margin}

Our final proposed model, \emph{sat-margin}, combines aspects from the previous two models.  It optimizes a sum of binary variables, like max-correct, but also aims to confidently predict each sample, like min-hinge.
\begin{align}
    & \max \sum_{k \in T} \sum_{j \in N_L}  q^k_j & \label{sat:obj}\\
    & (q^k_j = 1) \Longrightarrow (\hat{y}^k_j \cdot y^k_j \geq m) & \forall  j \in N_L, k \in T \label{sat:o_pos} \\
    & (q^k_j = 0) \Longrightarrow (\hat{y}^k_j \cdot y^k_j \leq m -\epsilon) & \forall  j \in N_L, k \in T \label{sat:o_neg} \\
    & q^k_j \in \{0,1\} & \forall j \in N_L, k \in T \label{sat:domain} \\
    & m = \frac{1}{2} \label{sat:margin}
\end{align}

The advantages of using the sat-margin model are that it tries to reach the same minimum objective value as min-hinge, without a need for defining PWL functions.  This is helpful when working with multiple objectives. Note that we use a margin of $m = \frac{1}{2}$. The margin could however be any value between $0$ and $1$.  With smaller margins, the model may be less confident in predictions but quicker to solve, while with larger margins the model must be more confident.

\subsection{Model compression}
\label{subsec:compress}

Besides optimal training of the NN weights, a further advantage of using a discrete optimization solver is that we can \emph{simultaneously} train the NN and optimize certain of its parameters.
Thus we next propose an extension to our MIP models that acts as a regularizer.  The purpose of this extension is to find the minimum number of neurons needed in the NN to fit to the training data.  The purpose is reflected in the objective \eqref{compress:obj}, where the sum of the binary variables $h_{\ell j}$ is minimized.  This binary variable is added for every neuron in the hidden layer(s) of the NN.  If the variable is $0$, the neuron's bias and all incoming and outgoing weights are set to be equal to $0$.  This effectively removes the corresponding neuron from the network.

This extension can be described as a method of model compression.  It can be added to our max-correct model and sat-margin model.  By adding this extension, our MIP models become multi-objective.  Note that as a consequence, the model compression cannot be effectively added to the min-hinge model as it relies on a PWL objective function; the Gurobi MIP solver cannot process a PWL objective(s) when in multi-objective mode.
\begin{align}
    & \min \sum_{\ell \in \mathcal{L}^{L-1}} \sum_{j \in N_\ell}  h_{\ell j} & \label{compress:obj}\\
    & h_{\ell j} = 0 \Longrightarrow
    \begin{cases}
    w_{i \ell j}  = 0 & \forall i \in N_{\ell-1}, j \in N_{\ell},   \\
    b_{\ell j}  = 0  & \hfill \ell \in \mathcal{L}^{L-1},  \\
    w_{j (\ell+1) m}  = 0 & \hfill m \in N_{\ell+1}
    \end{cases} \label{compress:params_zero}
\end{align}

\subsection{Batch training}
\label{subsec:batch}

Previous efforts with MIP models can feasibly train a NN with only a relatively low number of samples.  As our empirical results will demonstrate, our models described above improve upon the state of the art since they result in lower training runtimes and thus can handle more data.  Nevertheless, it would be beneficial to further increase the data that MIP models can accommodate.

We therefore introduce a novel batch training method that is inspired by gradient-based mini-batch training and ensemble training.  Stochastic gradient descent (SGD) with mini-batches does not train on all data at once; instead it iteratively updates parameters on small batches of data \cite{Goodfellow-et-al-2016}.  MIP models cannot simply update parameters for small batches.  Instead, we propose a method that resembles ensemble training.  Training data is  distributed randomly into batches.  A single MIP model is then trained on each batch.  This results in multiple MIP models that can be aggregated into a single model.  This is done for multiple iterations in an attempt to converge to a near-optimal model.

Each MIP model that is trained on a batch of data is independent.  Thus, we can exploit training them in parallel.  When aggregating models into a single model after training, we use the resulting networks' respective validation accuracy as the weight for a weighted average of the networks' parameters.  SGD generally trains over multiple iterations, or epochs.  To reach the best possible combined MIP model, we similarly train over multiple iterations.  After each iteration, we constrain the $P$ values (Equation~\ref{model:bound_domain}) for the next iteration's MIP models to be closer to the weighted average from the aggregated MIP model from the last iteration.  Before each iteration $\epsilon$, the lower bound of parameter $w_{i \ell j}$ is constrained to be the the maximum of $\{-P, w_{\ell i j} -P + \epsilon\}$.  Conversely, the upper bound of parameter $w_{i \ell j}$ is constrained to be the the minimum of $\{P, w_{\ell i j} + P - \epsilon\}$.  With this progressive constraining, we ensure that our algorithm will converge to the final weighted average after $P+1$ epochs.  Finally, as the converged model may not be optimal due to overfitting, we store each iteration's aggregated model and choose the model with the best validation accuracy afterwards.

The following steps summarize our proposed batch training algorithm: 
\begin{enumerate}
    \item Distribute all training data randomly into small batches (e.g., 100 samples each);
    \item Train individual MIP NN models on each batch in parallel;
    \item Combine all NNs into one NN using validation accuracies as weights for a weighted average;
    \item Constrain MIP NN parameter ranges for next epoch to be closer to weighted average;
    \item Repeat the above four steps until convergence.
    \item After convergence, choose the model from the iteration that had the highest validation accuracy.
\end{enumerate}
Detailed pseudocode can be found in Algorithm~\ref{alg:constrain}.  

\begin{algorithm*}[tb]
\caption{Progressive constraining batch training method for MIP models}
\begin{algorithmic}[1]
\State $W_{lb}[\ell,i,j], B_{lb}[\ell,j] \gets -P, \ \forall \ell \in \mathcal{L}, i \in N_{\ell - 1}, j \in N_\ell $ \Comment{Initialize lower bounds}
\State $W_{ub}[\ell,i,j], B_{ub}[\ell,j] \gets P, \ \forall \ell \in \mathcal{L}, i \in N_{\ell - 1}, j \in N_\ell$ \Comment{Initialize upper bounds}
\State $Weights \gets [\;], Biases \gets [\;], Vals \gets [\;]$
\For{e = 0,1,2, \dots, P} \Comment{Iterate over at most P+1 epochs}
\State $T \gets random.shuffle(T)$ \Comment{Shuffle all training data $T$ randomly}
\State $Batches \gets \{T_1, T_2, ..., T_D\}$ \Comment{Split training data into $D$ batches}
\State $Weights_e \gets [\;], Biases_e \gets [\;], Vals_e \gets [\;]$
\For{d = 1, \dots, D}
\State $model \gets NN(W_{lb},W_{ub},B_{lb},B_{ub})$ \Comment{Initialize NN model with bounds}
\State $W_d, B_d \gets model.train(data=T_d)$ \ {Optimize model for batch d}
\State $Val_d \gets calculate\_val\_acc(W_d, B_d)$ \Comment{Calculate validation accuracy}
\State $Weights_e.push(W_d)$;
$Biases_e.push(B_d)$;
$Vals_e.push(Val_d)$
\EndFor
\State $W_{avg} \gets weighted\_average(Weights_e, Vals_e)$
\State $B_{avg} \gets weighted\_average(Biases_e, Vals_e)$
\State $Val \gets calculate\_val\_acc(W_{avg}, B_{avg})$
\State $Weights.push(W_{avg})$;
$Biases.push(B_{avg})$;
$Vals.push(Val)$
\For{($\ell$,i,j) in $W_d$} \Comment{Iterate thro'
each connection (i,j) in every layer $\ell$ in $W_d$}
\State $W_{lb}[\ell,i,j] \gets max(-P,W_{avg}[\ell,i,j]-P+e)$ \Comment{
lower bounds of weights}
\State $W_{ub}[\ell,i,j] \gets min(P,W_{avg}[\ell,i,j]+P-e)$ \Comment{
upper bounds of weights}
\EndFor
\For{($\ell$,j) in $B_d$} \Comment{Iterate through each neuron j in every layer $\ell$ in $B_d$}
\State $B_{lb}[\ell,j] \gets max(-P,B_{avg}[\ell,j]-P+e)$ \Comment{Update lower bounds of biases}
\State $B_{ub}[\ell,j] \gets min(P,B_{avg}[\ell,j]+P-e)$\Comment{Update upper bounds of biases}
\EndFor
\EndFor
\State $i \gets argmax(Vals)$ \Comment{Find the index with the highest validation accuracy}
\State $W_{final} \gets Weights[i]$ \Comment{Use best index to extract best weights}
\State $B_{final} \gets Biases[i]$ \Comment{Use best index to extract best weights}
\end{algorithmic}
\label{alg:constrain}
\end{algorithm*}

\section{Experimental design}
\label{sec:expdesign}

We undertake four empirical experiments that train NNs using the proposed models.  Experiment~1 compares our MIP models to those of
Icarte \etal \cite{DBLP:conf/cp/IcarteICCMB19} on the standard MNIST dataset.
Experiment~2 compares training INNs using our three models to a gradient descent (GD) baseline on increasingly large training datasets.  Experiment~3 trains an INN using the sat-margin model while applying our model compression method.  Experiment~4 applies our proposed batch training method to utilize all available training data and compares to a SGD baseline.

Experiments 2A, 3 and 4~train NNs using the Adult dataset \cite{data:adult}.  The associated task
is binary classification.  There are 32,560 samples in the training set and 16,280 in the testing set.  Each sample represents an individual and the corresponding label denotes whether the individual has a yearly income of over 50K or not.  Each sample has 14 attributes, 8 of which are categorical and 6 are numerical. Experiment 2B trains NNs using the Heart dataset \cite{Heart}, which contains 303 samples in total. The associated task is also binary classification. Each sample in the Heart dataset has 14 attributes, 3 categorical and 11 continuous. Both datasets are pre-processed such that the categorical attributes are one-hot encoded and the numerical attributes are normalized to be between $0$ and~$1$.

Throughout we use Gurobi Optimizer version~9.0.1 \cite{Gurobi} to solve our MIP models.  Gurobi is a leading commercial MIP solver.  Gurobi parameters are set to their default values when optimizing.
All experiments except 2B
are run on an 8-core Intel Xeon Gold 6148 CPU at 2.40GHz with 32GB RAM.  Experiment~2B is run on a 4-core Intel i7-6600U CPU at 2.60GHz with 16GB RAM.

As our GD baseline, we use a method to train INNs introduced by Hubara \etal \cite{Courbariaux_Quantized}, using the squared hinge loss. The baseline uses the sign activation function and stochastic gradient descent.  Learning rates were tuned to reach the best performance for every experiment.  In Experiments~1
and~2,
a learning rate of \num{1e-2} was used.  In Experiment~4,
a learning rate of \num{1e-1} was found to be best.

The networks trained have one hidden layer containing 16 neurons.  We use the sign function as our activation function in the hidden layer.  Each network has two neurons in the final layer, one for each label the sample can take.  To assign a class to the sample, we choose the larger value of the output neurons.  If the values are equal for a sample, we choose the label by random tiebreak.

To shorten solving time, we do not require each model to be solved to optimality.  Reaching the global optimum with limited data will likely lead to over-training as well.  We therefore allow the MIP solver to stop optimizing once the network is ensured to have a training accuracy of above 90\%.

The source code is available: \url{https://github.com/tomasthorbjarnarson/mip_nn}

\subsection{Experiment 1}
\label{subsec:exp1}

The purpose
is to replicate the experiments of Icarte \etal \cite{DBLP:conf/cp/IcarteICCMB19}, training BNNs on the MNIST dataset.  Like in their experiments, we use a maximum runtime of 2 hours.  While their work
uses from 10 to 100 samples, we
train with 100 samples.  We compare our models to their min-weight and max-margin models.  We expect we
will reach higher accuracies in less time.

\subsection{Experiment 2A}
\label{subsec:exp2}

The purpose
is to study more thoroughly the feasibility of using MIP models to train NNs with limited data.
Each network is trained using up to 280 samples.  We investigate how our models perform compared to GD for such limited data.
We are interested in how the resulting networks generalize to the testing set.  We also consider how the runtimes of each model change with more training data.  We allow the solver a maximum runtime of 10 hours.

We also research the effects of increasing the range of the variables that represent weights and biases in the network.  We compare training BNNs where $P=1$ (Equation~\ref{model:bound_domain}), to training INNs with $P=3$, $7$ and $15$.  Each increase in range represents one extra bit needed to store a network's parameter in memory.

We hypothesize that the proposed MIP models will perform comparably to the GD baseline.  The min-hinge and sat-margin models should have similar testing accuracy but should both outperform the max-correct model.  However, we expect the max-correct model will have a much shorter runtime.

The increase in range for parameters should result in NNs that are easier to train.  Although the increased range results in larger search spaces, it will be easier to reach good solutions.  BNNs have very constrained variables and smaller search spaces. In contrast, the increased range of INNs could lead to fitting to training samples easier.  Because INNs may find better solutions quicker, we hypothesize that they will have shorter runtimes.  However, they may not generalize better.  Increased ranges of parameters could lead to more over-training, while BNNs are highly regularized.

\subsection{Experiment 2B}
\label{subsec:exp2B}

The purpose
is to research if similar findings are found on different datasets.  
This experiment is identical to Experiment~2A apart from we use the Heart dataset with up to 200 samples. This experiment is also run with a separate setup that has fewer cores and less memory than in Experiment~2B. We hypothesize that the results will follow a similar trend to the results from Experiment~2B. The sat-margin and min-hinge models should perform similarly to the GD baseline in terms of accuracy but will have longer runtimes.

\subsection{Experiment 3}
\label{subsec:exp3}

The next experiment assesses whether the network size can be optimised simultaneously with its training.
Specifically, we train a multi-objective INN MIP model, with the second objective function \eqref{compress:obj}.  We use the sat-margin MIP model and extend it with model compression.  Thus, we optimize both the sat-margin and the model compression objectives at once.  We apply a weight $\alpha$ to the model compression objective that denotes how much focus should be given to compressing the NN while training.  We compare values of $\alpha=0$ (no compression), $\alpha=0.1$ and $\alpha=0.01$. These $\alpha$ values are chosen to reflect that optimizing the sat-margin objective should be more important than compressing.

We train on 800 samples from the Adult dataset.  It is imperative to use as much training data as the model can handle within a time limit, as our model compression method finds the minimum number of neurons needed to fit to training data.  Thus, with less data, the resulting networks always reach the lower bound on the number of neurons in the hidden layer.  We set this lower bound to be two, as we have two neurons in the output layer.  We allow the solver a maximum runtime of 10 hours.

\subsection{Experiment 4}
\label{subsec:exp4}

Our final experiment applies our batch training methodology.  We again use the sat-margin INN MIP model and the Adult dataset.  We use 25,000 samples for the training dataset. The rest of the training data is used for validation.  We use a batch size of 100.  This results in training 250 NN MIP models for each epoch.  We assign 2 CPU cores per model, such that we can train 4 MIP models at once in parallel.  We then compare our results to a SGD baseline that similarly trains an INN with a batch size of 100.

\section{Results and discussion}
\label{sec:results}

\subsection{Empirical results}

Results for \textbf{Experiment~1} can be seen in Table~\ref{tab:extra_exp}.
Each MIP model is run five times.  Each run uses a random subset of the available data.  The table shows the average results over the runs.
On the standard MNIST dataset (which was used by Icarte \etal \cite{DBLP:conf/cp/IcarteICCMB19}), with a modest size of 100 samples, the two MIP models of Icarte \etal \cite{DBLP:conf/cp/IcarteICCMB19} each obtain a testing accuracy of 10\% by the time training is terminated at the time limit of 2 hours.  By contrast, all our methods have at least three times the testing accuracy. The fastest model, max-correct, completes in about 30 minutes of training.  Min-hinge, further, achieves a testing accuracy of over 50\%, easily beating also GD in terms of accuracy.

\begin{table*}[t]
\centering
\captionsetup{justification=centering}
\begin{tabular}{llll}
\toprule
& Training Accuracy \% & Testing Accuracy \% & Runtime [s]\\
\midrule
Max-correct & 97.8 $\pm$ 3.9 & 31.3 $\pm$ 2.3 & 2854.6 $\pm$ 1111.3 \\
Sat-margin & 69.2 $\pm$ 37.9 & 33.9 $\pm$ 14.1 & 7202.8 $\pm$ 2.4 \\
Min-hinge & 100.0 $\pm$ 0.0 & 51.1 $\pm$ 2.8 & 1852.1 $\pm$ 579.4 \\
\midrule
min\_w & 8.8 $\pm$ 3.1 & 10.3 $\pm$ 0.0 & 7200.0 $\pm$ 0.0 \\
max\_m & 8.8 $\pm$ 3.1 & 10.3 $\pm$ 0.0 & 7200.0 $\pm$ 0.0 \\
\midrule
gd\_nn & 70.0 $\pm$ 14.3 & 43.1 $\pm$ 7.2 & 998.9 $\pm$ 1173.3 \\

\bottomrule
\end{tabular}
\caption{Average results for Experiment 1 over five runs with standard deviation.}
\label{tab:extra_exp}
\end{table*}

\begin{figure}[t]
    \begin{subfigure}{0.5\columnwidth}
        \centering

        \includegraphics{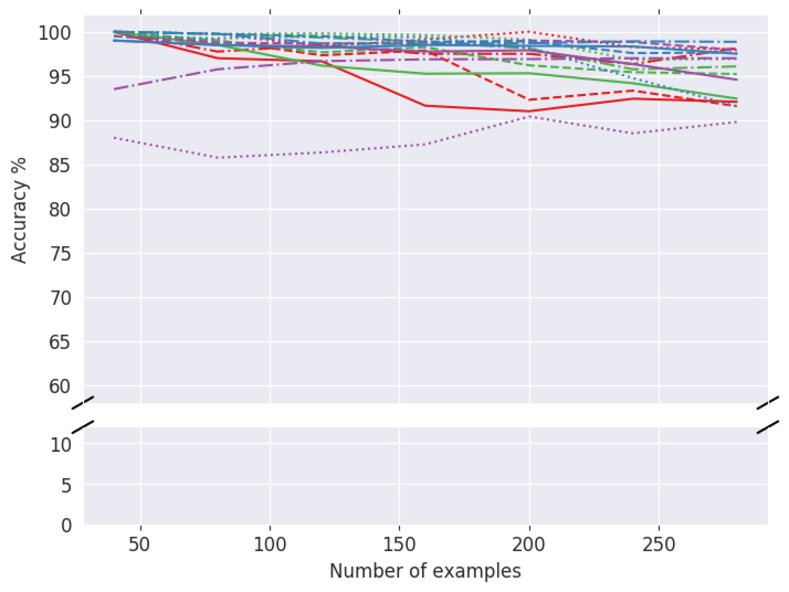}
        \caption{Training accuracy of models with different\\parameter ranges for Adult dataset.}
        \label{fig:exp1-train}
    \end{subfigure}
    \begin{subfigure}{0.5\columnwidth}
        \centering

        \includegraphics{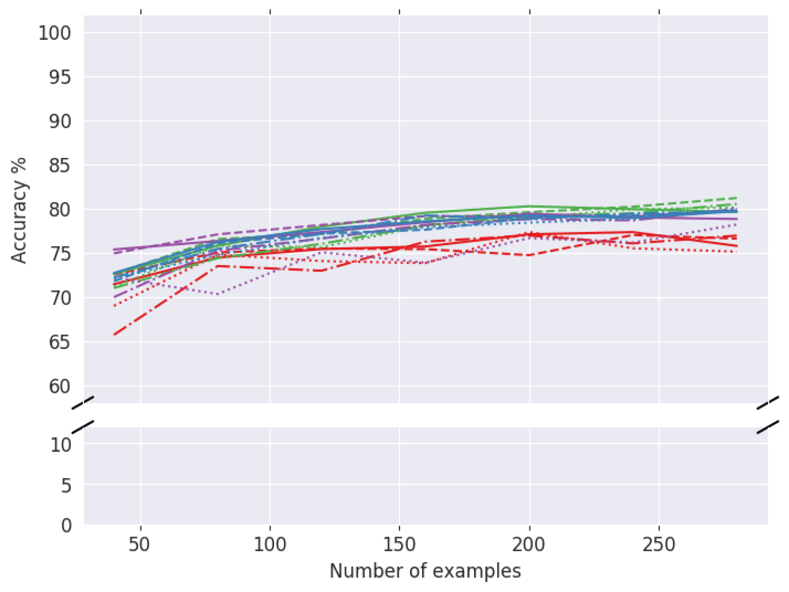}        
        \caption{Testing accuracy of models with different\\parameter ranges for Adult dataset}
        \label{fig:exp1-test}
    \end{subfigure}
    ~\\
    \begin{subfigure}{0.5\columnwidth}
        \centering

        \includegraphics{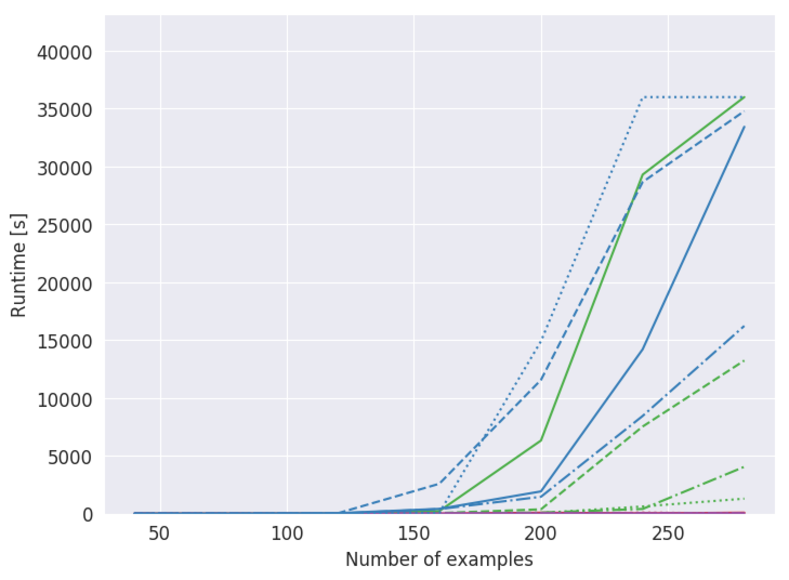}        
        \caption{Runtimes of models with different parameter ranges for Adult dataset}
        \label{fig:exp1-time}
    \end{subfigure}
    \begin{subfigure}{0.5\columnwidth}
        ~\\~\\~\\~~

        \includegraphics{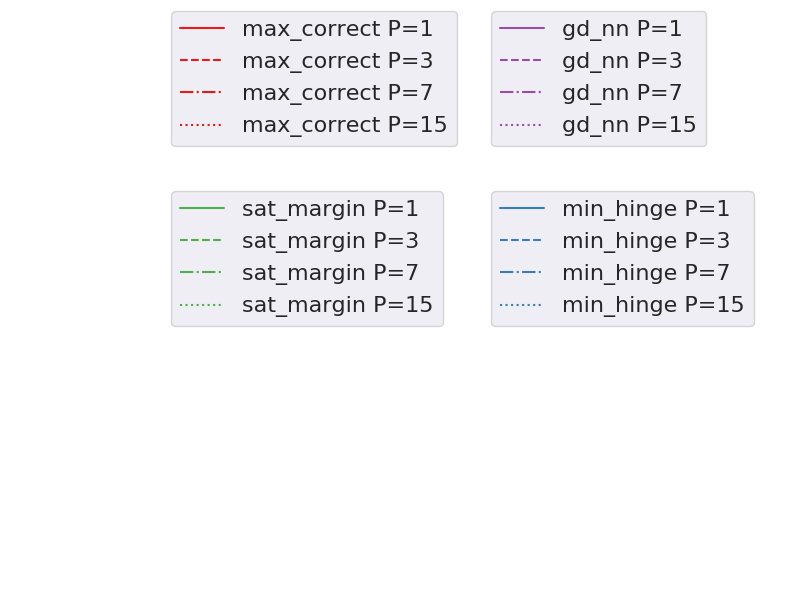}        
    \end{subfigure}
    \caption{Results for Experiment~2A (Adult dataset), averages over five runs.}
\end{figure}

Results for \textbf{Experiment~2A} can be seen in Figures~\ref{fig:exp1-train},~\ref{fig:exp1-test} and~\ref{fig:exp1-time}, which compare our models and the GD baseline (\emph{gd\_nn}).  For visual clarity, we do not show error margins in the figures.  

\begin{figure}[t]
    \begin{subfigure}{0.5\columnwidth}
        \centering

        \includegraphics{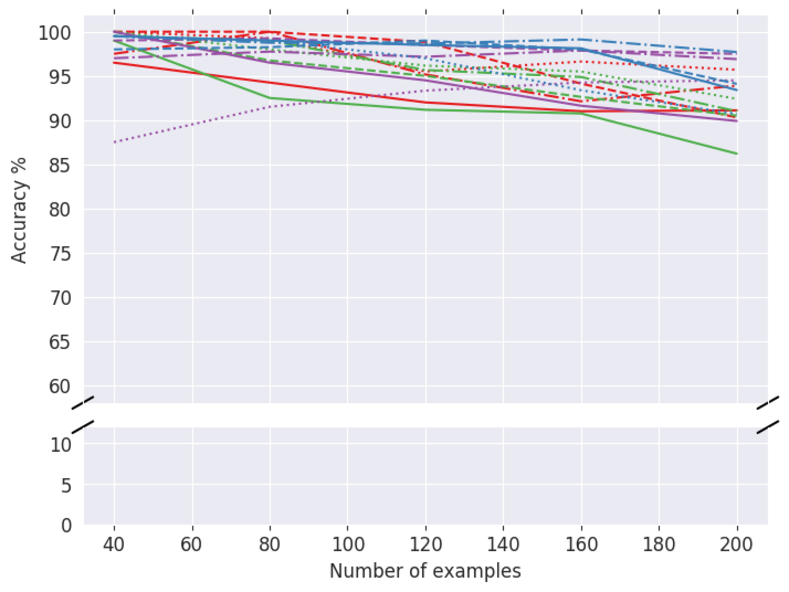}        
        \caption{Training accuracy of models with different\\parameter ranges for Heart dataset.}
        \label{fig:exp1-heart-train}
    \end{subfigure}
    \begin{subfigure}{0.5\columnwidth}
        \centering

        \includegraphics{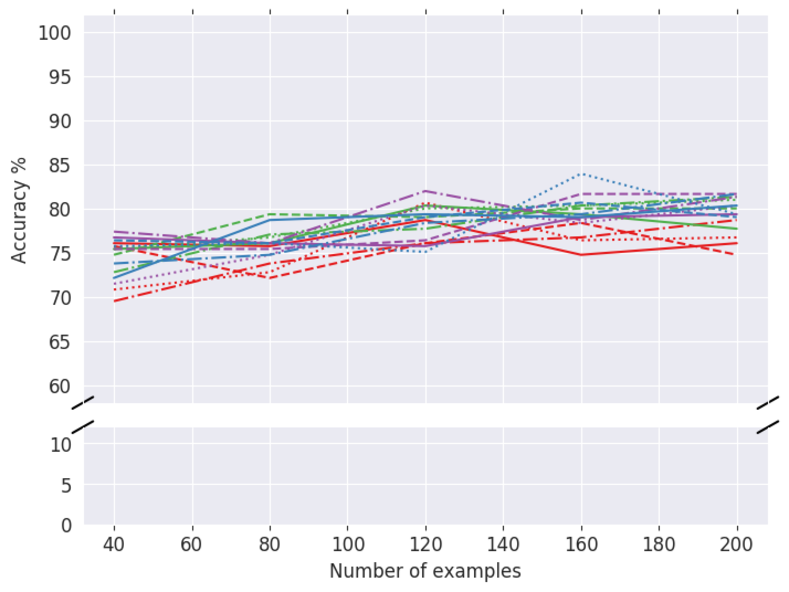}
        \caption{Testing accuracy of models with different\\parameter ranges for Heart dataset}
        \label{fig:exp1-heart-test}
    \end{subfigure}
    ~\\
    \begin{subfigure}{0.55\columnwidth}
        \centering

        \includegraphics{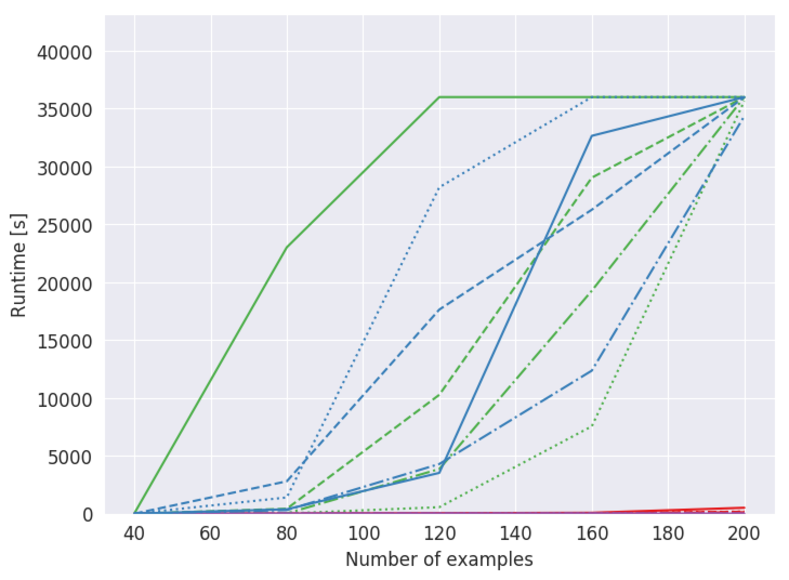}        
        \caption{Runtimes of models with different parameter ranges for Heart dataset}
        \label{fig:exp1-heart-time}
    \end{subfigure}
    \begin{subfigure}{0.5\columnwidth}
        ~\\~\\~\\~~

        \includegraphics{fig4}        
    \end{subfigure}
    \caption{Results for Experiment~2B (Heart dataset), averages over five runs.}
\end{figure}

Figure~\ref{fig:exp1-test} shows how well our models generalize and compare them to the GD baseline.  We see that with more data available, testing accuracy generally increases for every model.   The max-correct model is erratic and performs worst of the models.  The min-hinge and sat-margin models perform similarly and even slightly outperform GD, with increasing amounts of data.  On the other hand, we see that there is no big difference in testing accuracy when comparing different parameter ranges for each model.

Figure~\ref{fig:exp1-time} shows the evolution of runtimes with increasing amounts of data.  It becomes clear that with more than 200 samples to train on, the solving time for min-hinge and sat-margin drastically increases.  Max-correct and GD, on the other hand, have very short runtimes.  Here we see the biggest difference when comparing parameter ranges.  It is clear that for different $P$ values, runtimes can vary.  For the sat-margin model, there is a general trend that with larger ranges (i.e., $P=15$), the runtime is shorter than for constrained ranges (i.e., $P=1$).  For the min-hinge model however, the shortest runtime is found with $P=7$, while with $P=15$ the runtime is even longer than for the binary model.

Results for \textbf{Experiment~2B} can be seen in Figures~\ref{fig:exp1-heart-train},~\ref{fig:exp1-heart-test} and~\ref{fig:exp1-heart-time}. 
Figure~\ref{fig:exp1-heart-test} shows how well our models generalize and compares them to the GD baseline. Similarly to the results in Experiment~2A, the max-correct model performs worst of the models.  The min-hinge model, sat-margin model and GD baseline perform similarly, reaching 78--82\% accuracy with 200 training samples.

Figure~\ref{fig:exp1-heart-time} shows the evolution of runtimes with increasing amount of data. We again see great differences when using different $P$ values. For the sat-margin model, larger parameter ranges again always lead to decreased runtimes. For the min-hinge model, we again see that $P=7$ results in the shortest runtime. The max-correct model and GD again have very short runtimes in comparison to the min-hinge and sat-margin models.

\begin{table*}[t]
\centering
\captionsetup{justification=centering}
\begin{tabular}{llllll}
\toprule
& Training Acc. \% & Testing Acc. \% & Runtime [s]  & Neurons\\
\midrule
Sat-margin, $\alpha=0$   & 79.7 $\pm$ 3.63 & 78.1 $\pm$ 1.98 & 36000 $\pm$ 0 & 16.0 $\pm$ 0.0 \\
Sat-margin, $\alpha=0.1$  & 87.0 $\pm$ 2.01 & 78.8 $\pm$ 1.90 & 36000 $\pm$ 0 & 5.2 \ $\pm$ 5.42 \\
Sat-margin, $\alpha=0.01$  & 85.7 $\pm$ 2.91 & 77.9 $\pm$ 1.17 & 36000 $\pm$ 0 & 2.2 \ $\pm$ 0.40 \\
\bottomrule
\end{tabular}
\caption{Average results for Experiment~3 over five runs with standard deviation.}
\label{tab:exp2}
\end{table*}

Results for \textbf{Experiment~3} can be seen in Table~\ref{tab:exp2}.  We use the sat-margin model with $P=15$, since Experiment~2 found that it had a relatively short runtime while also reaching a good level of testing accuracy, when compared to the other models and parameter ranges.
Each MIP model is run five times for every $\alpha$ variation.  Each run uses a random subset with 800 samples from the available training data.  The table shows the average results over the runs.

When $\alpha=0$, no model compression is applied. Thus we see that the final number of neurons in the hidden layer is not reduced.  With $\alpha=0.1$ and $\alpha=0.01$, the models are compressed while training. All models reached 2 or 3 neurons in the hidden layer except for one run for $\alpha=0.1$, which did not manage to compress the model within the time limit.
All runs reach the time limit of 10 hours, however: thus on average the compressed models reach higher levels of training accuracy while also minimizing the size of the model.

Results for \textbf{Experiment~4} can be seen in Table~\ref{tab:batch}.  Again, the sat-margin model is used with $P=15$.  We compare our batch training method to SGD mini-batch training.  Both methods use a batch size of 100 samples and are run until convergence.  We run both methods five times and show the average results.
We see that the results for training and testing accuracies are very similar for our method and the SGD method.  Nevertheless, our method has a substantially longer runtime, as we discuss next.

\subsection{Discussion}
\label{subsec:discussion}

\begin{table*}[t]
\centering
\captionsetup{justification=centering}
\begin{tabular}{lllll}
\toprule
& Training Accuracy \% & Testing Accuracy \% & Runtime [s] \\
\midrule
Sat-margin   & 83.6 $\pm$ 0.18 & 83.8 $\pm$ 0.25 & 18900 $\pm$ 704  \\
SGD  & 85.6 $\pm$ 0.10 & 83.7 $\pm$ 0.13 & 319 \ \ \  $\pm$ 4.79  \\
\bottomrule
\end{tabular}
\caption{Average results for Experiment~4 over five runs with standard deviation.}
\label{tab:batch}
\end{table*}

Our proposed models clearly outperform the previous state of the art by Icarte \etal \cite{DBLP:conf/cp/IcarteICCMB19}.
As hypothesized, the max-correct model is quicker than our other proposed models.  However, because it does not require any confidence in predictions, it results in lower testing accuracies.  The min-hinge and sat-margin models perform similarly well as we had hoped.  They both push predictions to be more confident and therefore generalize better than max-correct.

We see that there is a considerable difference in runtime when training INNs compared to BNNs.  The increased range in parameters allows the sat-margin model to solve much quicker, without degrading generalization much.  This is not the case for min-hinge however.  The shortest runtime for min-hinge is found with value $P=7$, while $P=3$ and $P=15$ can have longer runtimes than with $P=1$.  This shows that while tuning the parameter range can greatly reduce runtime, there is no one range that will always perform best. 
For the max-correct model, there is little difference between runtimes and testing accuracy for different parameter ranges.  The sat-margin model is preferred to min-hinge and max-correct as it strives to be confident in predictions, and its runtime can be greatly reduced by using a larger parameter range than for a BNN.

Summarizing, there is a trade-off in increasing parameter ranges.  With larger ranges, more memory is needed to represent the network.  Nevertheless, with the aim of pushing the limits on how much data can be feasibly used to train on, training low-bitwidth INNs is preferable to training BNNs.

It is clear that model compression brings us even closer to our goal of using as much training data as possible.  Our method finds the minimum number of neurons needed in the network to fit to the training data.  As a side effect, large parts of the network are removed.  Thus, fewer parameters are optimized which leads to shorter runtimes.  The downside to our method is that by training with model compression we are likely underfitting to the training data.  The resulting network may be too simple and may therefore not generalize as well as possible.  This becomes clear by comparing testing accuracy in Experiment 1 for 280 samples in Figure~\ref{fig:exp1-test} to testing accuracy in Experiment 2 for 800 samples in Table~\ref{tab:exp2}.  Although the number of training samples is greatly increased, testing accuracy does not increase.

Lastly, our proposed batch training methodology succeeds in using all available training data.  Our results show that there is clear potential in applying batch training in combination with MIP.  Although our method is relatively simple, it manages to generalize as well as SGD.  Our method takes a considerably longer time to converge than SGD.  However, because it exploits parallelism, the runtime can be drastically reduced by distributing batches across more CPUs.

\section{Related work}
\label{sec:rw}

While an emphasis of work at the intersection of operations research and machine learning has been exploiting the latter to help solve optimization problems studied by the former, an important thrust is also the use of
optimization tools to advance the latter \cite{Bengio20:ml-co-survey}.  This is our purpose in this article.

Fischetti and Jo \cite{DBLP:journals/constraints/FischettiJ18} researched modelling NNs using MIP to optimize certain aspects of the network.  Instead of training using solvers, they use pre-trained networks and use solvers to find optimized adversarial examples.  They model the problem to modify examples minimally such as to fool the network into classifying the example incorrectly.
Tjeng \etal \cite{DBLP:conf/iclr/TjengXT19} go further into evaluating robustness of NNs with MIP by finding optimized adversarial examples.  They provide tight formulations for non-linearities in the models which result in considerably quicker solving times.
While Fischetti and Jo \cite{DBLP:journals/constraints/FischettiJ18} focus on multi-layer perceptrons as we do in this article, Tjeng \etal \cite{DBLP:conf/iclr/TjengXT19} examine the robustness of deeper networks as well as networks with convolutional and residual layers.

Anderson \etal \cite{DBLP:conf/ipco/AndersonHTV19} provide strong MIP formulations of pre-trained NNs.  Like 
Fischetti and Jo \cite{DBLP:journals/constraints/FischettiJ18} and Tjeng \etal \cite{DBLP:conf/iclr/TjengXT19}, they model ReLU networks and evaluate robustness of networks by modifying samples with minimal perturbations. However, their model removes the need for additional variables to model the ReLU function.  Anderson \etal \cite{Anderson20:strong} extend to max-pool operators.
Grimstad and Andersson \cite{Grimstad19:relu-surrogates} similarly optimize certain aspects of pre-trained NNs by using MIP: namely, using ReLU networks as surrogate models in MIP.  They highlight the importance of bound tightening techniques and how it effects the efficiency of the models.
The results show that ReLU networks are suitable as surrogate models in MIP, at least for small, shallow networks.
In contrast to these works, however, we directly train NNs using MIP.

The closest work to our is Icarte \etal \cite{DBLP:conf/cp/IcarteICCMB19}, who proceeded to directly train BNNs using MIP models.
Instead of optimizing a function that leads to high training accuracy, these authors introduce constraints that ensure the network fits to training data perfectly.  They then propose two variations of the model.  Variation 1 maximizes the number of zero-weight connections in the network, thus effectively removing as many unnecessary connections as possible.  Variation 2 maximizes margins on every neuron in the network, which should lead to more confident activations and predictions.
Our work differs 
in that our
models directly optimize loss functions to maximize accuracy.  Further, our models are more relaxed and therefore are more capable of handling more training data.  Our work is also more general, in that we are able to handle the important class of non-binary NNs.

Bah and Kurtz \cite{DBLP:journals/corr/abs-2007-03326} extend
Icarte \etal \cite{DBLP:conf/cp/IcarteICCMB19} with an improved MIP model, a local search algorithm and a two-stage robust optimization approach.  They do not treat non-binary NNs, nor compress models or batch train; they train on only 160 samples in 24 hours while we handle much more data.

Serra \etal \cite{Serra20:compression} use MIP to analyze an existing NN, identifying ReLUs with linear behaviour over the input,
which they replace using L1 regularization.
By contrast, our model compression determines,
in principle,
the optimal number of neurons during training. 

Bienstock \etal  \cite{DBLP:journals/corr/abs-1810-03218} provide a theoretical framework for training NNs using Linear Programming (LP).  This approach differs from ours as we focus on training with MIP, which allows a full capture of the NN semantics.  While our work focuses on fully connected NNs with only one layer, Bienstock \etal{} apply their methodology to deep NNs as well as convolutional NNs and Deep Residual Networks.  Because Bienstock \etal{} cannot accommodate integer constraints, they find an $\epsilon$-approximate solution, while our methodology searches for a globally optimal solution on the training dataset.

Ghosh and Datta Chaudhuri
\cite{Ghosh_Datta_Chaudhuri_2021} use methods of ensemble training that resemble the batch training method we apply in our work.  Note those authors look at stock prices, not MIP.  They apply random forests and bagging which both train multiple models and then classify with the use of majority voting from the models.  This differs to our method as we train multiple models before aggregating them into a single model which performs classification.  A voting scheme is not applicable to our batch training algorithm due to the iterative aggregation we apply.  However, it might be interesting for future work to train multiple MIP models on small batches of data and apply a voting scheme for classification.

Preitl et al.\@ \cite{Preitl06} present an approach to solving Multi Parametric Quadratic Programming (MPQP) models, but do not study training NNs.  Our min-hinge model is piecewise-linear as is solved in that paper.  The NNs in this article could also be modelled using quadratic programming.  That is not our focus, unlike authors who look at non-linear and global optimisation methods  \cite{Schurholt18,Schweidtmann19}.

\section{Conclusion}
\label{sec:conc}

This article advances a current idea of training neural networks using mixed integer programming solvers.  Such use of discrete optimization solvers for NN training represents an intriguing theoretical counterpoint to the standard approach of gradient descent training, but until our work has little practical value.

We demonstrated how to model and train integer-value neural networks, making three noteworthy further contributions.
First, we provide a framework that is flexible to change objective functions as well as the range of integers the network can encompass.  
Second, we show how to simultaneously optimize the number of neurons while training.  Third, we greatly increase the amount of data that can be processed by using batch training.

Empirical results demonstrate that our proposed models not only clearly outperform previous MIP models, but can also perform comparably to gradient descent baselines when using minimal data to train and relatively small NNs with integer or binary parameters.  
Further, solving time can be considerably improved by allowing NNs to have larger ranges for parameters, in comparison to binary parameters.
Moreover, training NNs using our methodology require minimal memory.  This can prove useful when deploying NNs to low-memory devices.

Training a NN using a single MIP model is still limited in some aspects.  In this article it is only found feasible to train NNs with a single hidden layer and relatively few neurons.  Training NNs with MIP is also limited to the amount of training data the MIP models can handle within a reasonable amount of time.
Nevertheless, our methods have raised the limits on the amounts of data  MIP models can handle.
Moreover, our batch training method counters this limitation by training several MIP models before combining them.
The batch training method
is simple and markedly effective.  Future work could ensure that batch training results in an optimal, converged network.  It would also be interesting to combine our model compression with batch training.  
Further, symmetry breaking in the MIP model could be usefully explored.

We investigated training integer-valued NNs using the sign activation function.  The NNs modelled in Grimstad and Andersson \cite{Grimstad19:relu-surrogates}, Fischetti and Jo \cite{DBLP:journals/constraints/FischettiJ18}, and latterly 
Anderson \etal \cite{Anderson20:strong} and Tsay \etal \cite{Tsay21:partition}
use the ReLU activation function.  Thus it would be interesting to extend our models to train using the ReLU function, as well as other potential activation functions; and to examine ideas from recent ReLU-training papers to varied activation functions.  While gradient-based training requires differentiable activation functions, this is not a requirement for MIP models.  This could lead to researching less explored, non-differentiable activation functions.

We compared our models to gradient-descent baselines on the Adult and Heart datasets.  The former dataset contains much more
data than a single MIP model can handle to date.  It would be interesting to use our models in combination with smaller, less explored datasets \cite{DBLP:journals/corr/abs-2007-03326}.  
We envisage the framework presented in this article being used in environments where minimal data is available to achieve a guaranteed optimized, low-memory NN that requires little to no hyper-parameter tuning.

{
\subsection*{Acknowledgements}

We thank the anonymous reviewers for their suggestions.
Thanks to E.~Demirovi\'{c}, S.~van der Laan, L.~Scavuzzo and A.~Schweidtmann.
This work was partially supported by TAILOR, a project funded by the EU Horizon 2020 research and innovation programme under grant number~952215.
}

%
%
%
%
%
%
%
%
%
%
%
%

\end{document}